\documentclass[runningheads]{llncs}
\usepackage{graphicx}
\usepackage{comment}
\usepackage{soul}
\usepackage{amsmath,amsfonts,amssymb} 
\usepackage[table]{xcolor}
\usepackage{enumitem}
\usepackage{multirow}
\usepackage{makecell}
\usepackage{accents}
\usepackage{upgreek}
\usepackage{bbm}

\usepackage{bm}
\usepackage{caption}
\usepackage[caption=false]{subfig}
\usepackage{mathtools}
\usepackage{booktabs}
\usepackage{algorithm}
\usepackage[noend]{algpseudocode}

\newcommand{\splitcond}[1]{%
\begin{tabular}[t]{@{}l@{}}#1\end{tabular}%
}

\newcommand*{\Scale}[2][4]{\scalebox{#1}{$#2$}}
\newcolumntype{M}[1]{>{\centering\arraybackslash}m{#1}}

\newcommand{\Break}{\State \textbf{break}}

\algrenewcommand\alglinenumber[1]{\tiny #1:}
\algrenewcommand\algorithmicindent{0.9em}

\usepackage[accsupp]{axessibility} 

\usepackage[pagebackref,breaklinks,colorlinks]{hyperref}

\usepackage[capitalize]{cleveref}
\crefname{figure}{Fig.}{Figs.}
\crefname{section}{Sec.}{Secs.}
\crefname{table}{Tab.}{Tabs.}
\crefname{algorithm}{Alg.}{Algs.}
\crefname{equation}{Eq.}{Eqs.}

\graphicspath{{figs/}}

\begin{document}
\pagestyle{headings}
\mainmatter
\def\ECCVSubNumber{10} 

\title{Look at Adjacent Frames: Video Anomaly Detection without Offline Training} 


\titlerunning{Look at Adjacent Frames}
%
\author{Yuqi Ouyang \and
Guodong Shen \and
Victor Sanchez}
\authorrunning{Y. Ouyang et al.}
%
\institute{University of Warwick, UK\\
\email{\{yuqi.ouyang, guodong.shen, v.f.sanchez-silva\}@warwick.ac.uk}\\
}
\maketitle

\begin{abstract}
We propose a solution to detect anomalous events in videos without the need to train a model offline. Specifically, our solution is based on a randomly-initialized multilayer perceptron that is optimized online to reconstruct video frames, pixel-by-pixel, from their frequency information. Based on the information shifts between adjacent frames, an incremental learner is used to update parameters of the multilayer perceptron after observing each frame, thus allowing to detect anomalous events along the video stream. Traditional solutions that require no offline training are limited to operating on videos with only a few abnormal frames. Our solution breaks this limit and achieves strong performance on benchmark datasets.
\end{abstract}

\section{Introduction}
\begin{figure}[!t]
\begin{minipage}[t]{0.645\linewidth}
\centering
\includegraphics[scale=0.233]{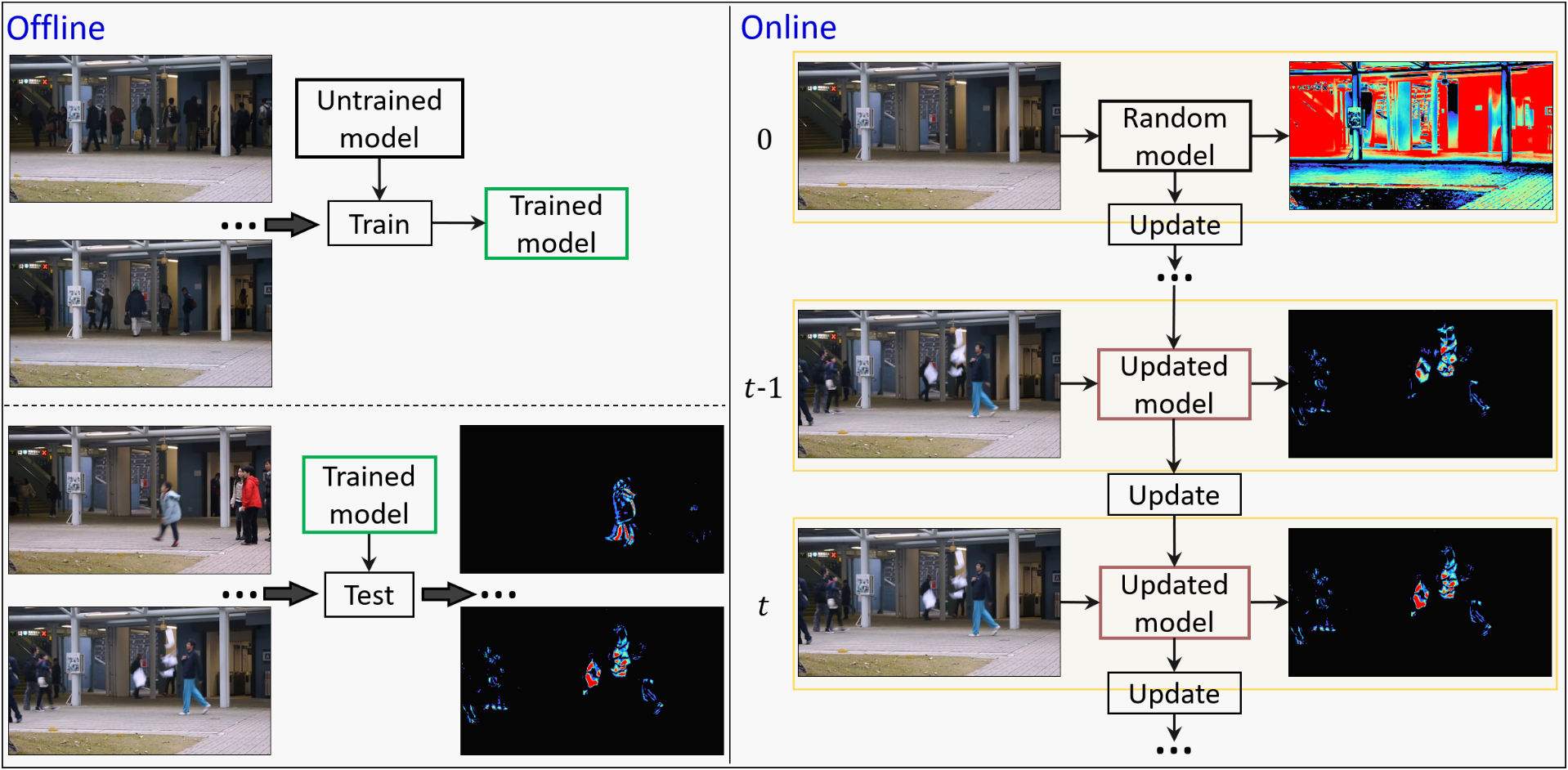}
\caption{\small{Working mechanisms of offline and online VAD.}}
\label{fig:VADwp}
\end{minipage}
\hfill
\begin{minipage}[t]{0.346\linewidth}
\centering
\includegraphics[scale=0.2387]{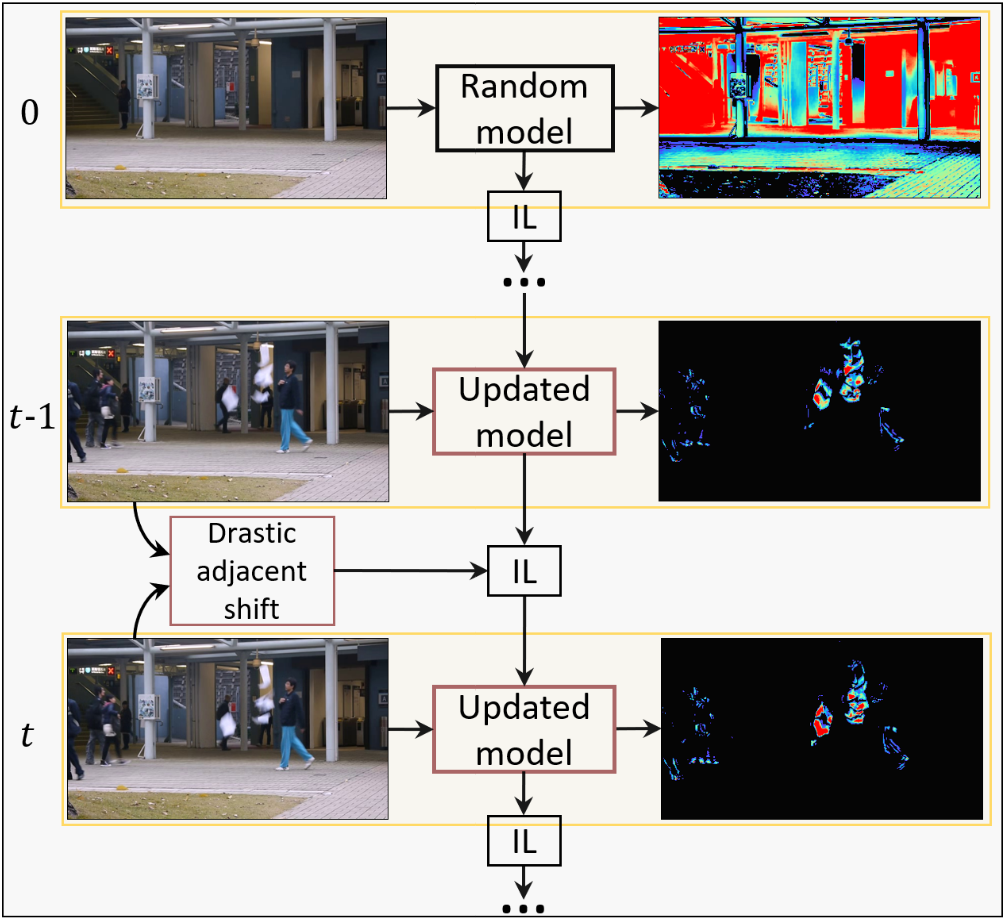}
\caption{\small{Idea of online VAD.}}
\label{fig:VADonline}
\end{minipage}
\vspace{-2pt}
\end{figure}

Video anomaly detection (VAD) aims at detecting anomalous events in a video scene. Since it is hard to define all possible anomalous events a priori and moreover, these anomalies may occur infrequently, VAD is rarely solved by supervised learning. It is then common to exclusively rely on normal video data to train a model for the detection of anomalous events. Hence, the lack of examples of abnormal events during training defines the inherent challenges of this task. However, real-life experiences show that a person could react to a biker moving among pedestrians because the biker moves distinctively, without knowing that the biker is an abnormality in this context. Enlightened by such human intelligence, VAD can also be solved without any knowledge learned from an offline training process by analyzing spatio-temporal differences between adjacent frames since these differences are usually significant where anomalous events occur \cite{Shuffle,UnMasking,MC2ST}. The working mechanism of this type of VAD solution, which requires no offline training and is hereinafter referred to as online VAD\footnote{\scriptsize{Note that in this paper, online VAD does not refer to VAD solutions that are trained offline but operate at high frame rates.}}, is illustrated in \cref{fig:VADwp} along with the working mechanism of offline VAD. Note that instead of training the model offline, online VAD updates the model sequentially after observing each video frame, thus requiring no training data.

One main advantage of online VAD is that a model can learn beyond the data used for offline training and adapt much better to the data source during operation, which eases the issue of concept drift caused by, e.g., any data distribution differences between the offline and online data. In offline VAD, a model is trained first to understand the patterns of videos depicting normal events. Once trained, it is used on new videos depicting normal events and possibly, abnormal events. In this case, the concept drift due to the distribution differences between the offline and online data depicting normal events may result in poor performance during operation. To deal with such a concept drift, one can use human support to regularly identify those normal frames on which the model performs poorly, and then retrain the model to recognize these frames correctly. However, such an approach undoubtedly comes with an extra workload.

Despite the fact that the performance of VAD solutions has improved recently \cite{SSMTL,CRF}, online VAD solutions are still scarce \cite{SurveyIVAD,SurveyVAD}. Moreover, the existing ones are usually flawed. For example, \cite{Shuffle,UnMasking,MC2ST}, which are pioneers in online VAD, attain poor performance on videos with a large number of anomalous events. Motivated by these limitations, we present our idea for online VAD in \cref{fig:VADonline}. Our idea focuses on the spatio-temporal differences between adjacent frames, hereinafter referred to as adjacent shifts. It uses an incremental learner (IL) that accounts for the adjacent shifts and sequentially updates the model from a set of randomly-initialized parameters. The IL is expected to easily adapt to gradual adjacent shifts, which usually exist between normal frames and should result in no anomaly detections. Conversely, the IL is expected to encounter difficulties in adapting to drastic adjacent shifts, which usually exist between abnormal frames or between a normal and an abnormal frame, thus resulting in anomaly detections. More specifically, as illustrated in \cref{fig:Solution}, our solution generates error maps as the detection results by reconstructing spatio-temporal information and pixel coordinates into pixel values and comparing the reconstructed frames with the original ones. In this work, we use the discrete wavelet transform (DWT) to summarize the spatio-temporal information of a video sequence and a Multi-layer Perceptron (MLP) for the frame reconstruction task. The MLP is set to be updated by the IL while adapting to the adjacent shifts, after observing each frame. The contributions of this work are summarized as follows:

\begin{figure}[!t]
\begin{minipage}[t]{0.602\linewidth}
\centering
\includegraphics[scale=0.217]{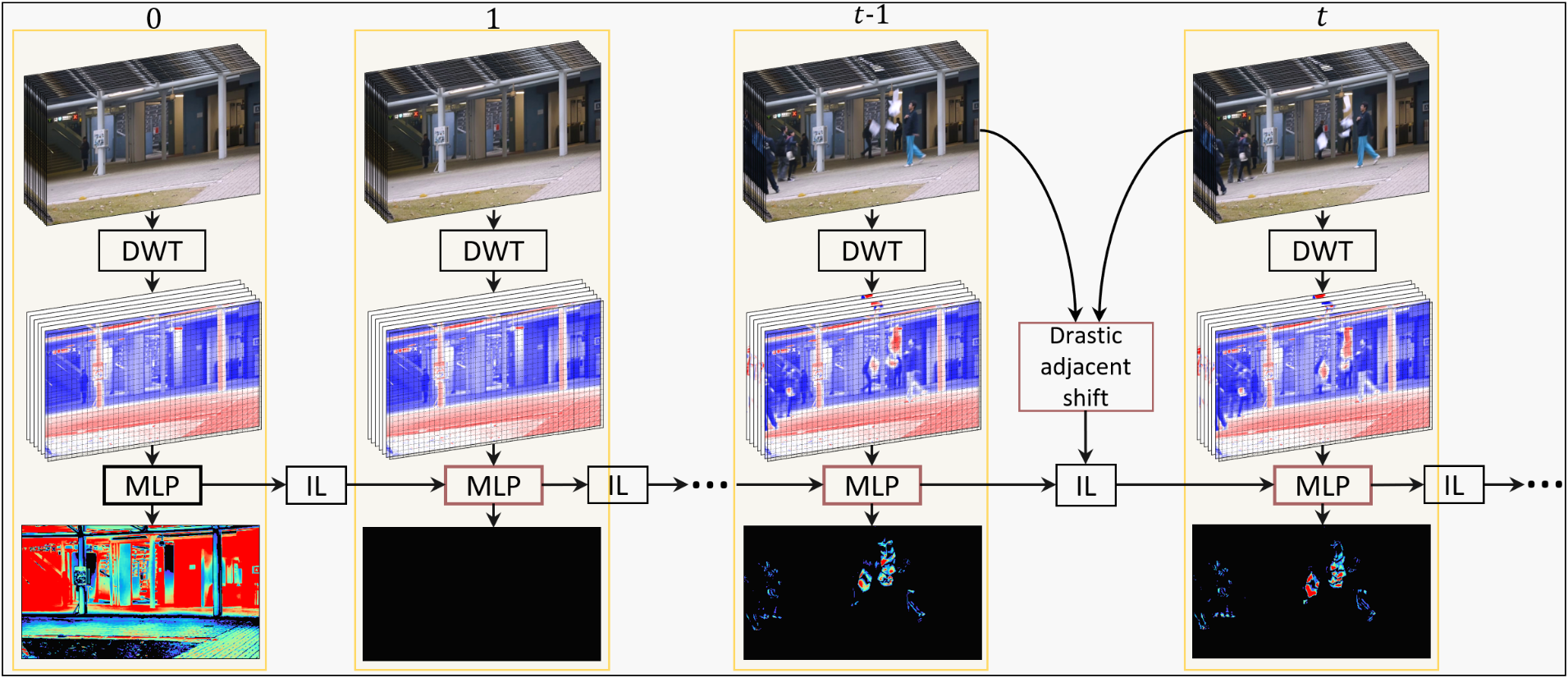}
\caption{\small{Our solution for online VAD.}}
\label{fig:Solution}
\end{minipage}
\hfill
\begin{minipage}[t]{0.388\linewidth}
\centering
\includegraphics[scale=0.177]{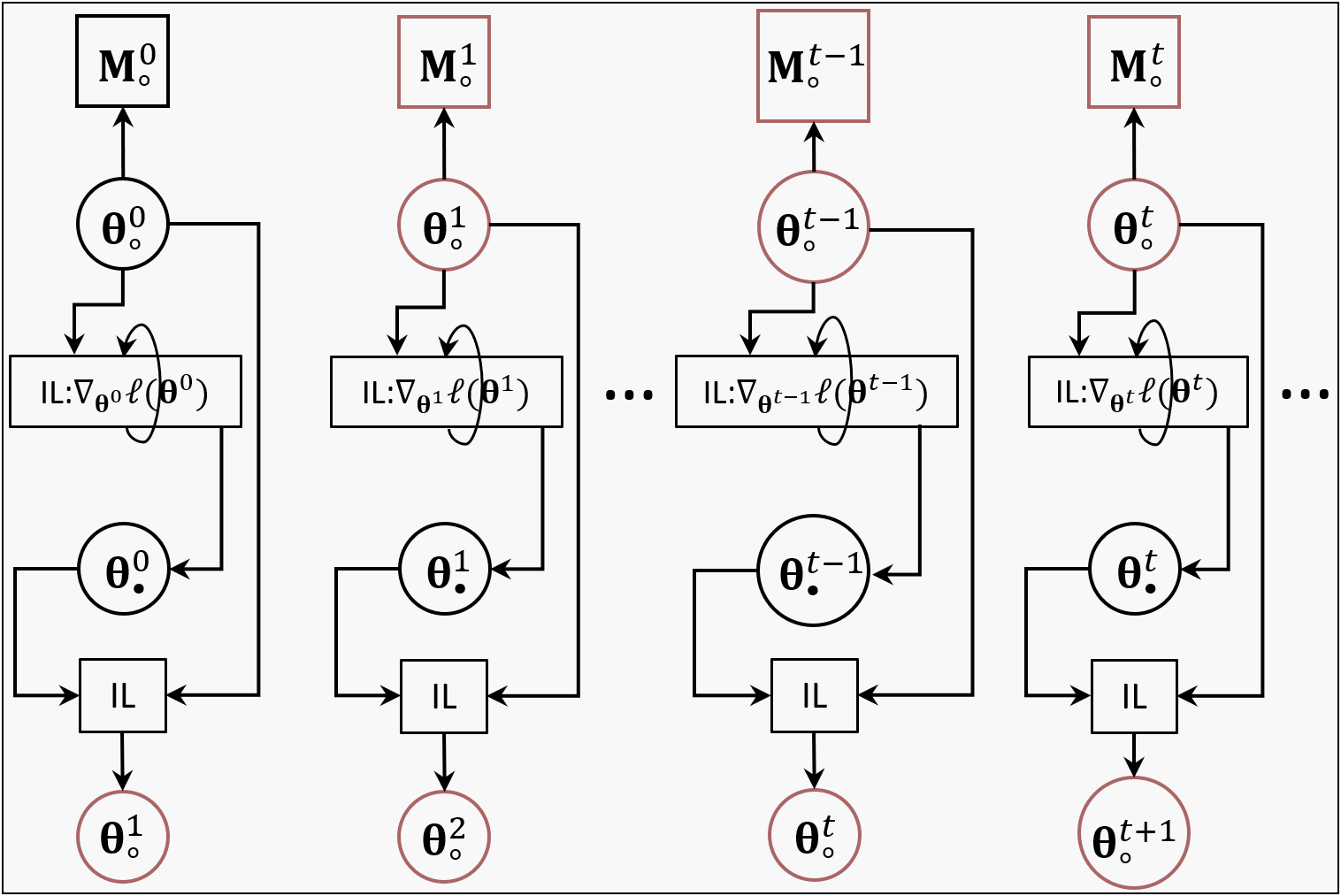}
\caption{\small{Our problem definition.}}
\label{fig:Problem}
\end{minipage}
\vspace{-2pt}
\end{figure}

\begin{itemize}[noitemsep,nolistsep]
\item We introduce the first MLP-based model that uses frequency information to produce pixel-level VAD results.
\item More importantly, we design a novel solution for online VAD, i.e., detecting anomalous events with \textbf{no} offline training, where the network parameters are optimized sequentially after observing each frame starting from random initialization.
\item We achieve state-of-the-art performance on benchmark datasets regardless of the temporal order or the number of abnormal frames in a video.
\end{itemize}

\section{Related Work} \label{RelatedWork}
\noindent \textbf{Offline VAD.}
State-of-the-art offline VAD solutions use deep-learning techniques. These models can be classified as unsupervised, weakly-supervised, or self-supervised depending on the training data and how these data are used. Unsupervised models exclusively employ normal videos for training, with data mapping via an encoder-decoder structure being the most frequently used strategy. The baseline approaches in this context are based on data reconstruction or prediction, where anomalous events are detected based on the reconstruction or prediction errors, respectively \cite{Conv-AE,STAE,TSC,Skeleton,AnoPCN,HF-VAD,Guodong1}. Adversarial losses can also be applied in such approaches, where the generator is adversarially trained to perform the data mapping so that abnormal videos lead to unrealistic outputs with large reconstruction or prediction errors \cite{GANs,FFP,MLAD,UnetGAN,Gold}. If the features extracted from the normal and abnormal videos form two distinct and compact clusters, anomalous events can be detected in a latent subspace.
In this type of approach, two ways are usually used to train a model. Either separately training the feature extraction and clustering steps \cite{AMDN,FRCN-Action,Objc,SACR,GMM-DAE}, or jointly training both steps end-to-end \cite{LSA,MemAE,MemAE2020,Clustering,SAGC,AMMC-Net}. 

Weakly-supervised models use a small number of abnormal frames with labels for training. Based on the triplet loss \cite{MLEP} or multiple instance learning strategy \cite{UCF-Crime,MAF,GCLNC,MNIST,RTFM,CRF}, such models can learn to increase the inter-class distance between normal and abnormal data. Weakly-supervised models outperform unsupervised models but they require abnormal data with ground truth labels \cite{CLAWS,MNIST,RTFM}. 

Self-supervised models are becoming increasingly popular \cite{Two-Stage,SDOR}. These models are trained with both normal and abnormal videos but without ground truth labels. They are usually trained iteratively by simulating labels through decision-making and then optimizing parameters based on the simulated labels. Data collection is easier for these models because they do not require ground truth labels. However, because these models may use some of the test videos in the training stage, their results should be interpreted with caution.

\noindent \textbf{Online VAD.}
Compared to offline VAD solutions, online models are limited both in number and performance \cite{SurveyIVAD,SurveyVAD}. Three main models \cite{Shuffle,UnMasking,MC2ST} have been designed to tackle VAD online. The seminal work in \cite{Shuffle} detects anomalous events by aggregating anomaly values from several frame shuffles, where the anomaly value in each shuffle is calculated by measuring the similarity between all previous frames and the frames in a sliding window. Although that particular model cannot work with video streams, it sets an important precedent for VAD online. The follow-up works in \cite{UnMasking,MC2ST} propose analyzing adjacent frame batches, where anomalies are defined as abrupt differences in spatio-temporal features between two adjacent batches. However, instead of randomly initializing a model and continuously optimizing it along the video stream, these two works repeat the random initialization of parameters every time a new frame is observed, ignoring the fact that frames may share common information, such as the scene background. Moreover, the performance of these three solutions, i.e., \cite{Shuffle,UnMasking,MC2ST}, degrades when used on videos with a large number of abnormal frames.\footnote{\scriptsize{Videos with more than \emph{50}\% of frames being abnormal.}} The reason for this is that those models assume a video only contains a few abnormal frames, thus a large number of abnormal frames violate their assumptions regarding the spatio-temporal distinctiveness of abnormal frames in a video.

\noindent \textbf{Offline VAD with further optimization.}
Recently, several offline VAD solutions that are further optimized online have been proposed \cite{VAD-CL,MUF,NOLA}. Specifically, these models are first trained offline and then refined online to reduce false positives. Namely, the model processes each frame online and adds potential normal frames to a learning set where false positives identified manually are also included \cite{VAD-CL}. Such a learning set is then used to further optimize the model so that it can more accurately learn the normal patterns beyond the offline training data. Scene adaptation has also been recently addressed in VAD \cite{r-GAN,MPN} based on recent advance in meta-learning \cite{MAML}. Specifically, the model is first trained offline to produce acceptable results for a variety of scenes, and is then further optimized with only a few frames from a specific scene of interest, thus rapidly adapting to that scene.

\section{Proposed Solution}\label{sec:Model}
\noindent \textbf{Problem definition.}
Based on our objective of generating pixel-level detections online in the form of error maps, we start by providing the problem definition:

\begin{itemize}[noitemsep,nolistsep]
\item At current timestep $t$, based on the network parameters $\bm{\uptheta}_{\circ}^{t}$, compute the error map $\mathbf{M}_{\circ}^{t}$, where anomalies are indicated at the pixel-level. 
\item Define the loss $\Scale[1.2]{\ell}(\Scale[1]{\bm{\uptheta}}^{t})$ to optimize $\bm{\uptheta}_{\circ}^{t}$ into $\bm{\uptheta}_{\bullet}^{t}$.
\item Use $\bm{\uptheta}_{\circ}^{t}$ and $\bm{\uptheta}_{\bullet}^{t}$ to define $\bm{\uptheta}_{\circ}^{t+1}$ for computing an error map for timestep $t$\,+\,$1$.
\end{itemize}

Note that we use two types of subscripts, i.e., a clear circle $\circ$ and a black circle $\bullet$, where $\circ$ ($\bullet$) denotes the initial (final) network parameters or the detection results before (after) the optimization process. We illustrate our problem definition in \cref{fig:Problem} starting from the first frame of a video stream, i.e., $t$\,=\,$0$, where the IL is in charge of updating the network parameters from $\bm{\uptheta}_{\circ}^{t}$ to $\bm{\uptheta}_{\bullet}^{t}$ and subsequently computing $\bm{\uptheta}_{\circ}^{t+1}$.

\begin{figure}[!t]
\centering
\includegraphics[scale=0.365]{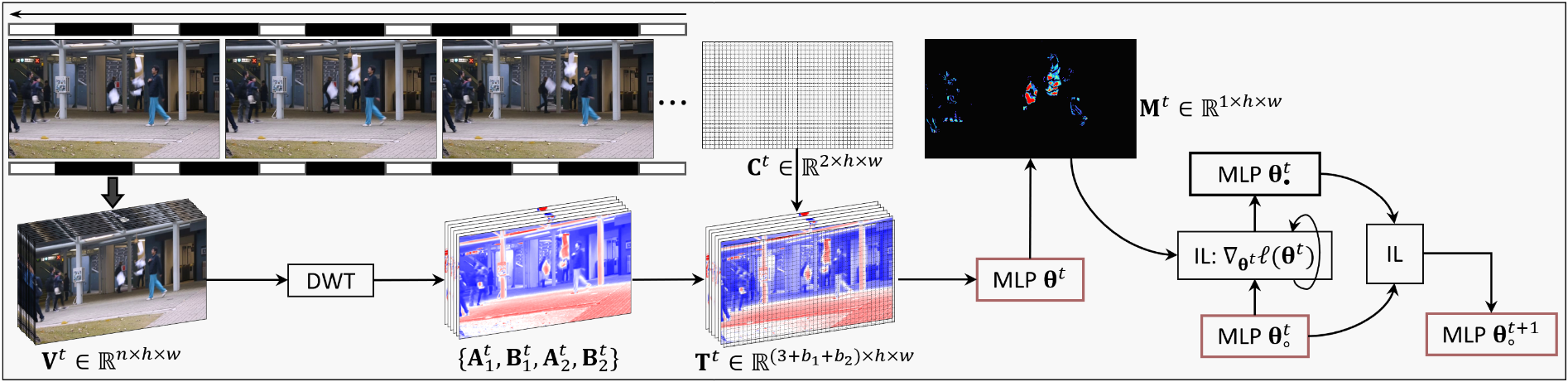}
\caption{\small{The workflow of our solution at timestep $t$. $b_1$ and $b_2$ respectively indicate the number of high-frequency maps in tensors $\mathbf{B}_{1}^{t}$ and $\mathbf{B}_{2}^{t}$ of the temporal DWT.}}
\label{fig:Workflow}
\end{figure}

\subsection{Workflow}
\begin{figure}[!t]
\begin{minipage}[t]{0.494\linewidth}
\centering
\includegraphics[scale=0.179]{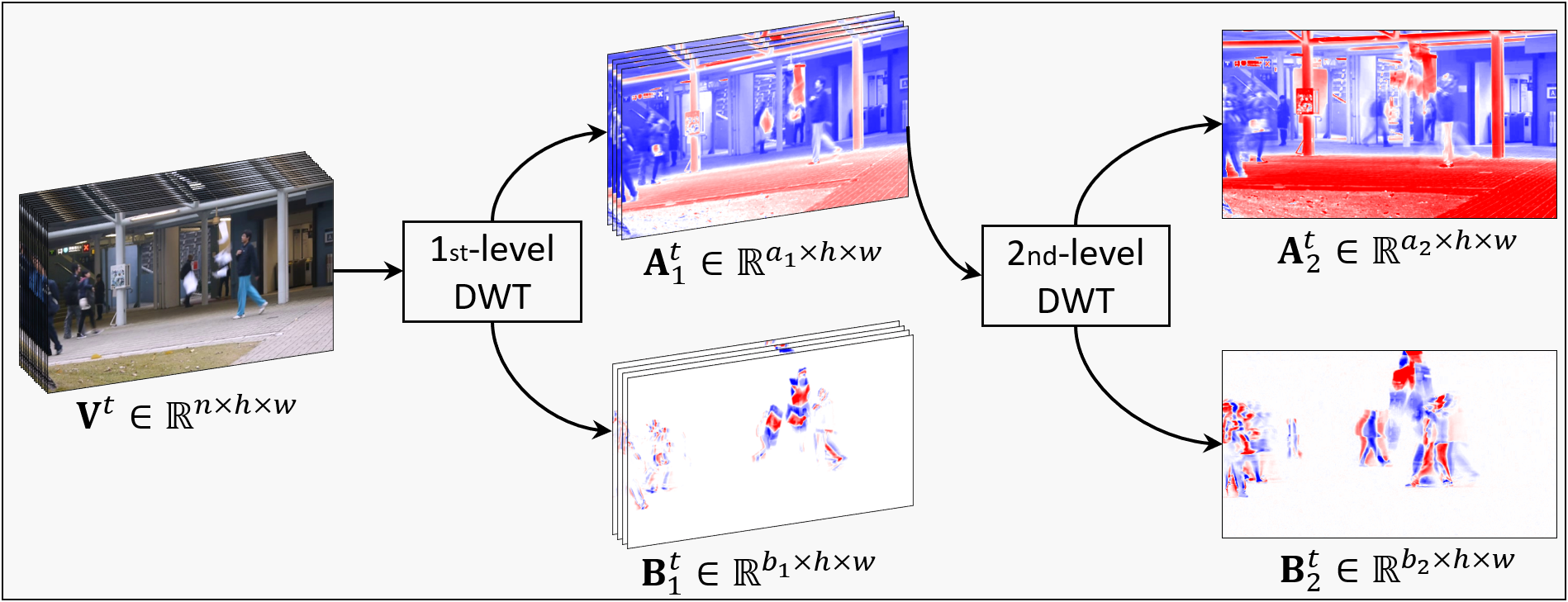}
\end{minipage}
\hfill
\begin{minipage}[t]{0.496\linewidth}
\centering
\includegraphics[scale=0.179]{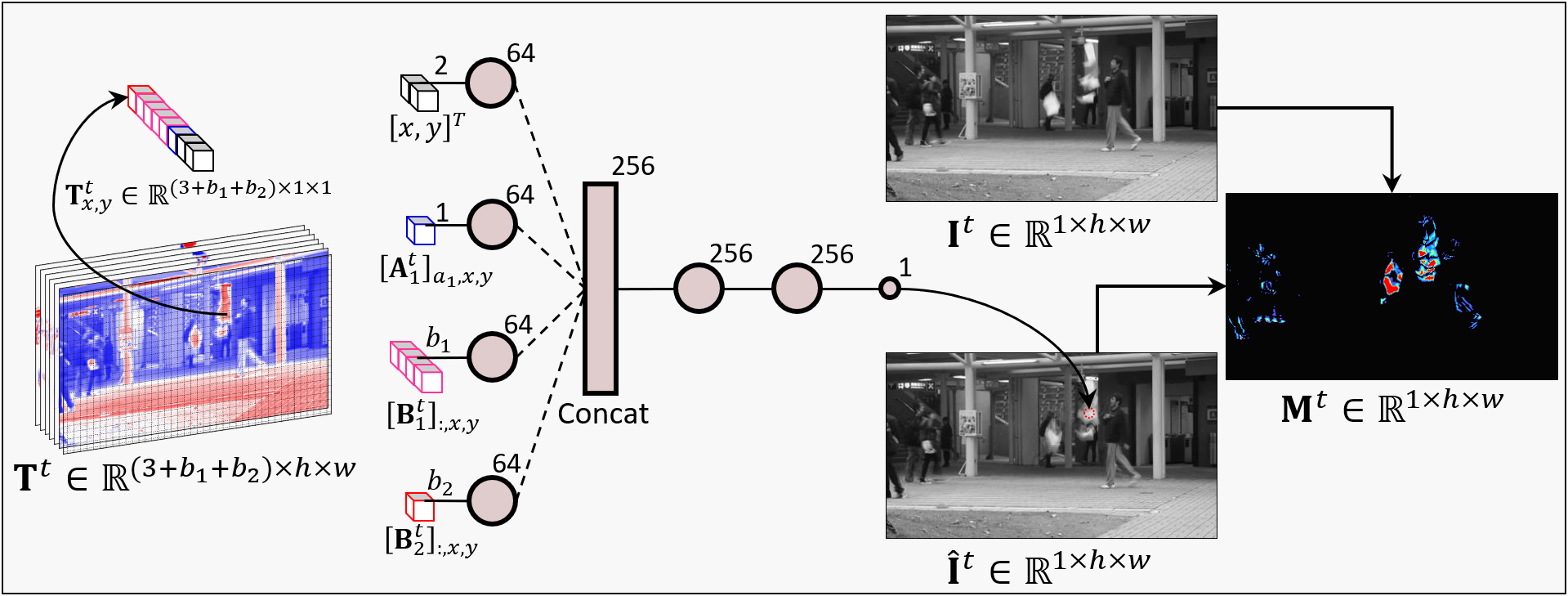}
\end{minipage}
\caption{\small{Details of our temporal DWT (left) and our MLP (right). $a_1$ and $a_2$ ($b_1$ and $b_2$) respectively indicate numbers of low-frequency (high-frequency) maps in tensors $\mathbf{A}_{1}^{t}$ and $\mathbf{A}_{2}^{t}$ ($\mathbf{B}_{1}^{t}$ and $\mathbf{B}_{2}^{t}$) of the temporal DWT. Layers of the MLP are depicted with the dimensions of their outputs. The error map is computed by comparing two grayscale images.}}
\label{fig:DWTandMLP}
\end{figure}

\cref{fig:Workflow} illustrates the workflow of our solution at the current timestep $t$.\footnote{\scriptsize{In \cref{fig:Workflow,fig:DWTandMLP}, the channel dimensions of the data are omitted.}} It first applies a temporal DWT to a set of $n$ frames of size $h \times w$, denoted by $\mathbf{{V}}^{t}$, to generate DWT coefficients, where the last frame of $\mathbf{{V}}^{t}$ is the current frame $\mathbf{I}^{t}$. The DWT coefficients are used in conjunction with the pixel coordinates $\mathbf{C}^{t}$ as the input $\mathbf{T}^{t}$ of a pixel-level MLP to reconstruct $\mathbf{I}^{t}$, thus leading to an error map $\mathbf{M}^{t}$ by comparing the reconstruction results with the ground truth frame. $\mathbf{M}^{t}$ is then used by the IL to define the loss $\Scale[1.2]{\ell}(\Scale[1]{\bm{\uptheta}}^{t})$. Based on $\Scale[1.2]{\ell}(\Scale[1]{\bm{\uptheta}}^{t})$, the IL updates the network parameters from $\bm{\uptheta}_{\circ}^{t}$ to $\bm{\uptheta}_{\bullet}^{t}$, and then calculates $\bm{\uptheta}_{\circ}^{t+1}$ for the next timestep.

\noindent \textbf{Discrete wavelet transform.}
In this work, we use the temporal DWT to summarize the spatio-temporal information of a video sequence because it has been shown to provide motion information that fits the human visual system \cite{DWT}. As shown in \cref{fig:DWTandMLP} (left), two levels of temporal DWT are performed on the set of frames $\mathbf{V}^{t}$. The temporal DWT results in four tensors of sub-bands. Specifically, two low-frequency tensors, $\mathbf{A}_{1}^{t}$ and $\mathbf{A}_{2}^{t}$, and two high-frequency tensors, $\mathbf{B}_{1}^{t}$ and $\mathbf{B}_{2}^{t}$. Note that the temporal DWT does not change the spatial dimensions. The input tensor $\mathbf{T}^{t}$ of our MLP is defined as:

\begin{equation}
\centering
\mathbf{T}^{t} =\left[\mathbf{C}^{t}, \,\mathbf{A}^{t}, \,\mathbf{B}_{1}^{t}, \,\mathbf{B}_{2}^{t}\right],
\label{eq:InputForMLP}
\end{equation}
\normalsize
where $\mathbf{A}^{t}$$\,\in\,$$\mathbb{R}^{1\times{h}\times{w}}$ is the last low-frequency map of the tensor $\mathbf{A}_{1}^{t}$. Here $\mathbf{T}^{t}$ is formed by stacking the coordinate tensor $\mathbf{C}^{t}$, the low-frequency map $\mathbf{A}^{t}$ (appearance information), the first-level high-frequency tensor $\mathbf{B}_{1}^{t}$ (sparse motion information) and the second-level high-frequency tensor $\mathbf{B}_{2}^{t}$ (dense motion information) along the first dimension.

\noindent \textbf{Multilayer perceptron.}
We use an MLP to reconstruct frames because it has been shown to effectively map pixel coordinates into pixel values \cite{Sin,FourierFeature,Meta-NRI}. As illustrated in \cref{fig:DWTandMLP} (right), our MLP maps $\mathbf{T}^{t}$ into the reconstructed frame $\hat{\mathbf{{I}}}^{t}$. Given such an MLP with parameters $\bm{\uptheta}^{t}$, the resulting error map, $\mathbf{M}^{t}$, and its mean squared error (MSE), $\Scale[1]{\upepsilon}^{t}$, are respectively computed as follows:

\begin{equation}
\centering
\mathbf{M}^{t} = \big(\mathbf{I}^{t}-{f}_{\Scale[0.55]{\bm{\uptheta}}^{\Scale[0.45]{t}}}(\mathbf{T}^{t})\big)^{\odot^{2}},
\label{eq:ErrMap}
\end{equation}
\normalsize

\begin{equation}
\centering
\Scale[1]{\mathrm{\upepsilon}}^{t} = \frac{1}{hw}{\big\Vert \mathbf{M}^{t} \big\Vert}_{1},
\label{eq:MSE}
\end{equation}
\normalsize
where ${f}_{\Scale[0.55]{\bm{\uptheta}}^{t}}(\cdot)$ denotes the mapping function of the MLP, $\odot^{2}$ indicates element-wise square, and ${\Vert\cdot\Vert}_{1}$ indicates ${\ell}_1$-norm.\footnote{\scriptsize{To compute the error map and the MSE when optimizing the MLP, please follow \cref{eq:ErrMap,eq:MSE} but replace $\bm{\uptheta}^{t}$ with current parameter set involved in the optimization; e.g., $\mathbf{M}_{\circ}^{t} = \big(\mathbf{I}^{t}-{f}_{\Scale[0.55]{\bm{\uptheta}}_{\Scale[0.55]{\circ}}^{\Scale[0.45]{t}}}(\mathbf{T}^{t})\big)^{\odot^{2}}$, $\Scale[1]{\mathrm{\upepsilon}}_{\circ}^{t} = \frac{1}{hw}{\big\Vert \mathbf{M}_{\circ}^{t} \big\Vert}_{1}$.}} The MLP is optimized by an IL, as detailed next.

\subsection{Incremental Learner}
As illustrated in \cref{fig:ILoneTime} (left), at the current timestep $t$, the IL comprises a comparator that defines the loss $\Scale[1.2]{\ell}(\Scale[1]{\bm{\uptheta}}^{t})$ from the MLP results, an adapter that optimizes the parameters of the MLP based on $\Scale[1.2]{\ell}(\Scale[1]{\bm{\uptheta}}^{t})$, and a clipper that calculates the initial parameters of the MLP for the next timestep, i.e., $\bm{\uptheta}_{\circ}^{t+1}$.

\noindent \textbf{The comparator.}
This component compares the reconstruction results of frame $\mathbf{I}^{t-1}$, as reconstructed by the MLP with parameters $\bm{\uptheta}_{\bullet}^{t-1}$, with those of frame $\mathbf{I}^{t}$, as reconstructed by the MLP with parameters $\bm{\uptheta}^{t}$. Based on this comparison, it defines the loss $\Scale[1.2]{\ell}(\Scale[1]{\bm{\uptheta}}^{t})$ for the MLP at timestep $t$. This loss is based on MSE values:

\vspace{-10pt}
\begin{equation}
\centering
\Scale[1.2]{\ell}(\bm{\uptheta}^{t}) = \mathrm{Relu}(\Scale[1]{\upepsilon}^{t}-\Scale[1]{\upepsilon}_{\bullet}^{t-1}),\,\,\,\,\, \splitcond{if\,\,\emph{t}\,$\geqslant$\,1}.
\label{eq:ILcommonLoss}
\end{equation}
\normalsize

The rationale behind the loss in \cref{eq:ILcommonLoss} is that such comparison accounts for the adjacent shift between frames $\mathbf{I}^{t-1}$ and $\mathbf{I}^{t}$. \cref{fig:ILoneTime} (top-right) shows an example, where a drastic adjacent shift allows a well-fitted MLP ($\bm{\uptheta}_{\bullet}^{t-1}$) to accurately reconstruct frame $\mathbf{I}^{t-1}$, while preventing the precedent MLP ($\bm{\uptheta}_{\circ}^{t}$) from accurately reconstructing frame $\mathbf{I}^{t}$. Hence, \cref{eq:ILcommonLoss} implicitly measures the adjacent shift by resulting in a large loss value, 
i.e., large $\Scale[1.2]{\ell}(\bm{\uptheta}_{\circ}^{t})$.\footnote{\scriptsize{To compute the loss when optimizing the MLP, please follow \cref{eq:ILcommonLoss} or \cref{eq:ILstartLoss} but replace $\bm{\uptheta}^{t}$ with current parameter set involved in the optimization; e.g., $\Scale[1.2]{\ell}(\bm{\uptheta}_{\circ}^{t}) = \mathrm{Relu}(\Scale[1]{\upepsilon}_{\circ}^{t}-\Scale[1]{\upepsilon}_{\bullet}^{t-1}),$ (if $t$\,$\geqslant$\,$1$).}}

\begin{table}[!t]
\begin{minipage}[!b]{0.565\linewidth}
\centering
\includegraphics[scale=0.322]{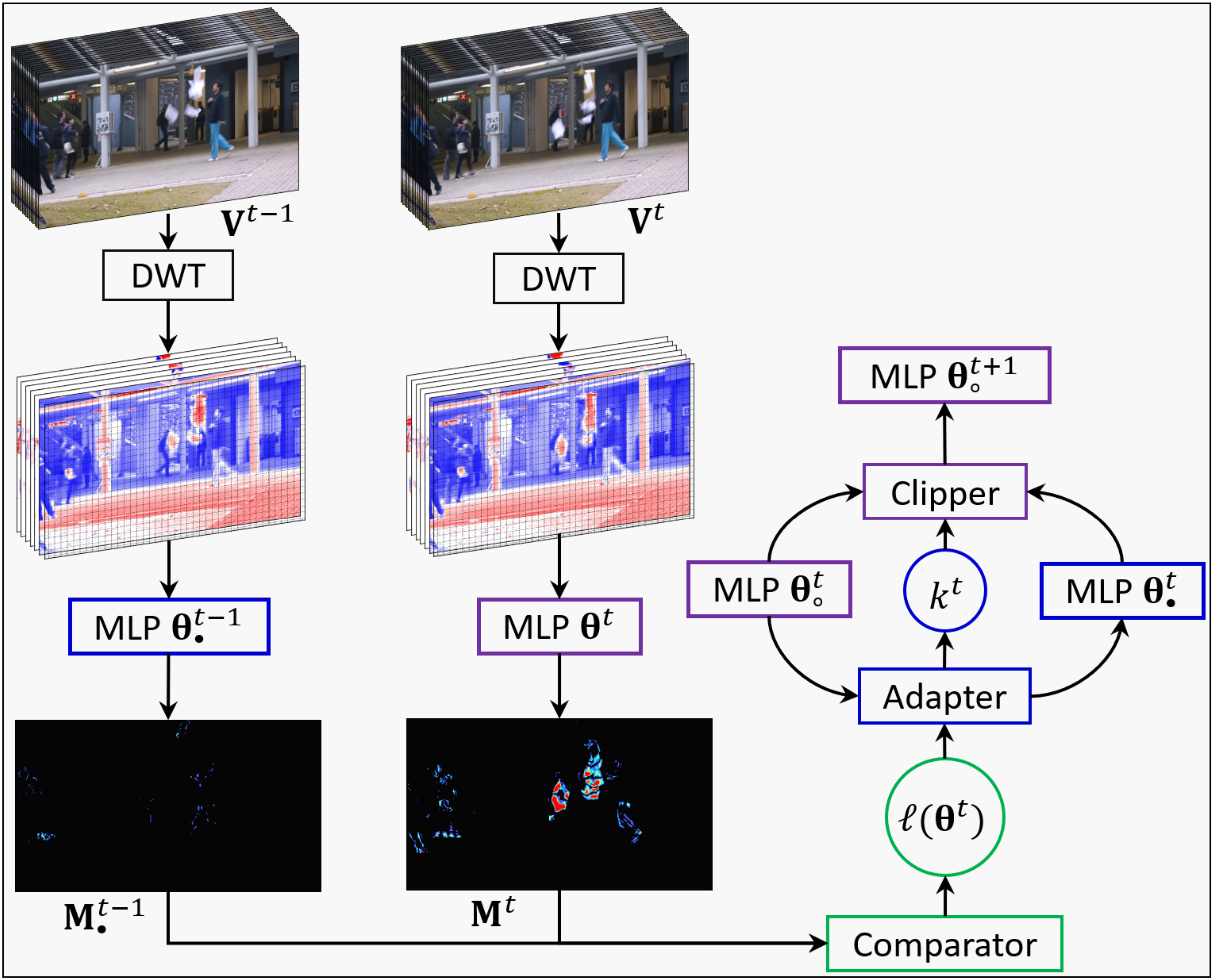}
\end{minipage}
\hfill
\begin{minipage}[!b]{0.425\linewidth}
\centering
\includegraphics[scale=0.257]{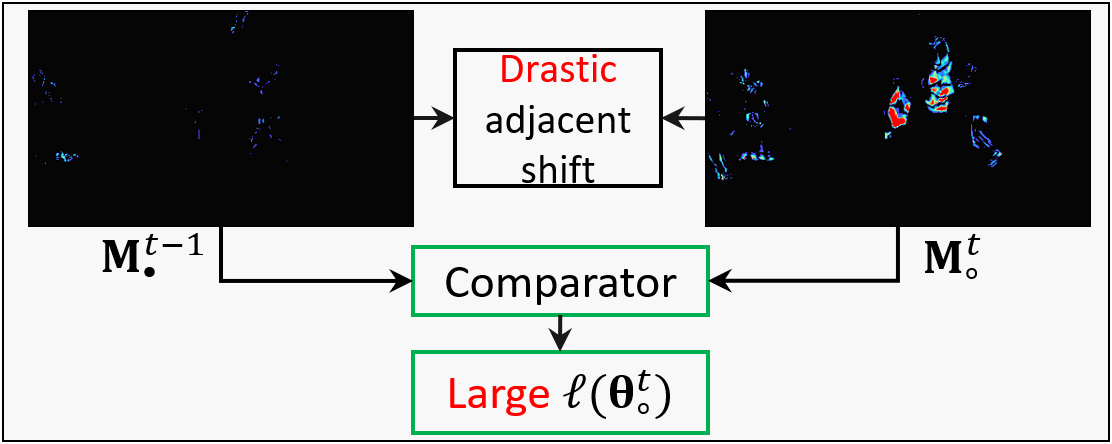}
\vfill
\includegraphics[scale=0.257]{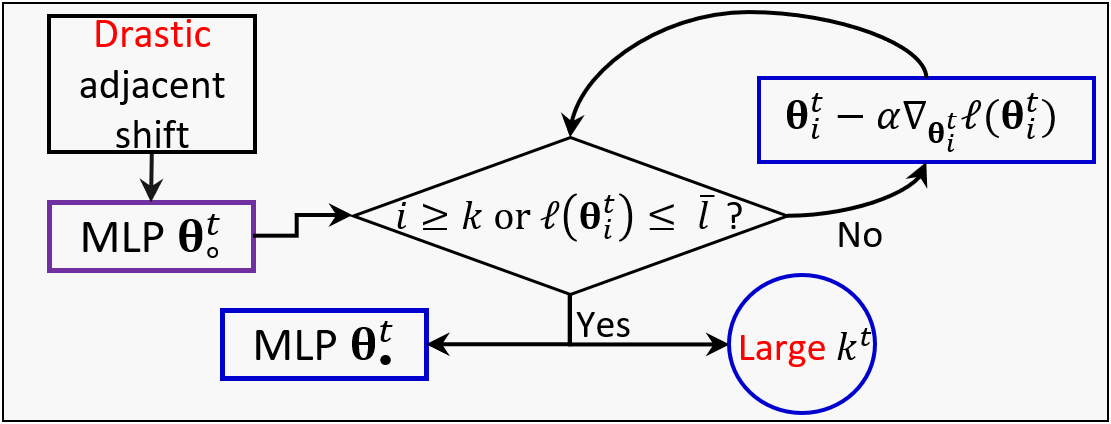}
\vfill
\includegraphics[scale=0.257]{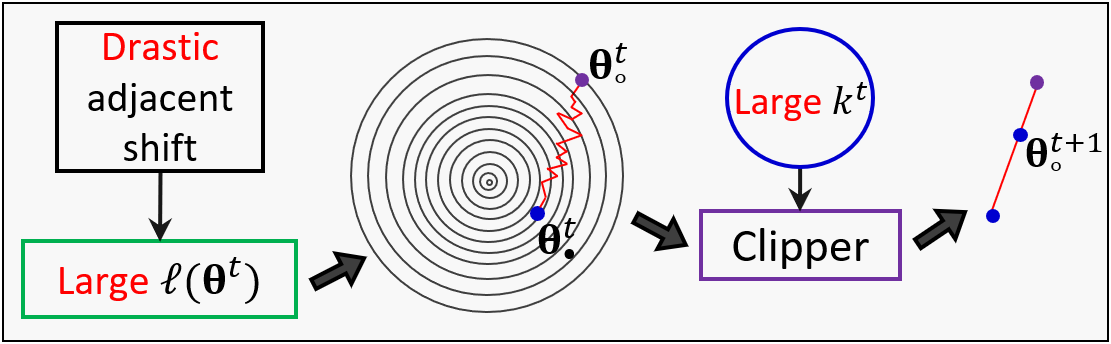}
\end{minipage}
\vspace{3pt}
\captionof{figure}{\small{The incremental learner at the current timestep $t$ (left). An example of a drastic adjacent shift that results in a large value for the loss function (top-right). The iterative process used to update the parameters of the MLP (middle-right). Computation of the initial parameters of the MLP for timestep $t$\,+\,$1$ (bottom-right).}}
\label{fig:ILoneTime}
\vspace{-15pt}
\end{table}
\normalsize

For the first frame of a video sequence, i.e., timestep $t$\,=\,$0$, note that there is no previous information for the comparator to define the loss in \cref{eq:ILcommonLoss}. We say that, in this case, the randomly-initialized MLP is under a cold start and may not generate accurate reconstruction results. Hence, we adjust \cref{eq:ILcommonLoss} for $t$\,=\,$0$ as follows:

\vspace{-10pt}
\begin{equation}
\centering
\Scale[1.2]{\ell}(\Scale[1]{\bm{\uptheta}}^{t}) = \mathrm{Relu}(\Scale[1]{\upepsilon}^{t}-\overline{\Scale[1]{\upepsilon}}),\,\,\,\,\, \splitcond{if\,\,\emph{t}\,$=$\,0},
\label{eq:ILstartLoss}
\end{equation}
\normalsize
where $\overline{\Scale[1]{\upepsilon}}$ is a user-defined MSE value that defines a target value to be achieved for an accurate reconstruction.\footnote{\scriptsize{We set $\Scale[1]{\upepsilon}_{\bullet}^{t-1} = \max\big(\Scale[1]{\upepsilon}_{\bullet}^{t-1}, \overline{\Scale[1.0]{\upepsilon}}\big)$ in \cref{eq:ILcommonLoss} to prevent our MLP from being overfitted towards a MSE value lower than the lower-bound $\overline{\Scale[1.0]{\upepsilon}}$.}}

\noindent \textbf{The adapter.}
Based on the loss $\Scale[1.2]{\ell}(\Scale[1]{\bm{\uptheta}}^{t})$, the adapter adapts to the adjacent shift by optimizing the parameters of the MLP over iterations of gradient descent (GD), the $(i$\,+\,$1)_{\mathrm{th}}$ GD iteration is defined as:

\begin{equation}
\centering
\bm{\uptheta}_{i}^{t} \xrightarrow{-\nabla_{\bm{\uptheta}_{i}^{t}}\ell(\bm{\uptheta}_{i}^{t}), \,\,\, \mathrm{if}\,\, \ell(\bm{\uptheta}_{i}^{t}) \,>\, \overline{l} \,\,\mathrm{and}\,\, i \,<\, \overline{k}} \bm{\uptheta}_{i+1}^{t},
\label{eq:ILoneGD}
\end{equation}
\normalsize
where $i$$\,\in\,$$[0,\overline{k}]$. Here we use two user-defined parameters, i.e., $\overline{k}$ and $\overline{l}$, to control the end of the iterative optimization: the adapter stops at the current GD iteration if the number of GD iterations is large enough ($i$\,$\geqslant$\,$\overline{k}$) or if the loss is small enough ($\Scale[1.2]{\ell}(\Scale[1]{\bm{\uptheta}}_{i}^{t})$\,$\leqslant$\,$\overline{l}$). This iterative process is illustrated in \cref{fig:ILoneTime} (middle-right), where the adapter optimizes the parameters from $\bm{\uptheta}_{\circ}^{t}$ to $\bm{\uptheta}_{\bullet}^{t}$.\footnote{\scriptsize{Note that $\bm{\uptheta}_{\circ}^{t}$\,=\,$\bm{\uptheta}_{\Scale[0.65]{i=0}}^{t}$.}}

Let $k^{t}$ denote the number of iterations of GD used by the adapter at timestep $t$. A large loss $\Scale[1.2]{\ell}(\Scale[1]{\bm{\uptheta}}^{t})$, potentially caused by the drastic adjacent shift, will force the adapter to spend many iterations of GD, i.e., a large $k^{t}$, to optimize the network parameters.

\noindent \textbf{The clipper.}
According to \cref{fig:ILoneTime} (left), the loss $\Scale[1.2]{\ell}(\bm{\uptheta}^{t+1})$ at the \underline{next} timestep $t$\,+\,$1$, is computed by comparing $\mathbf{M}_{\bullet}^{t}$, as computed under the parameter set $\bm{\uptheta}_{\bullet}^{t}$, and $\mathbf{M}^{t+1}$, as computed under the parameter set $\bm{\uptheta}^{t+1}$. Since the adapter generates $\bm{\uptheta}_{\bullet}^{t}$, we need to define the initial parameter set $\bm{\uptheta}_{\circ}^{t+1}$. This is accomplished by the clipper. 

As illustrated in \cref{fig:ILoneTime} (middle-right), a large value of $k^{t}$, i.e., the number of iterations of GD used by the adapter at timestep $t$, may imply a drastic shift between frames $\mathbf{I}^{t-1}$ and $\mathbf{I}^{t}$. Such a drastic shift may indicate that the current frame $\mathbf{I}^{t}$ is abnormal. Hence, the MLP may have been well-fitted to reconstruct the abnormal frame $\mathbf{I}^{t}$, and such a well-fitted parameter set $\bm{\uptheta}_{\bullet}^{t}$ may not be appropriate to initialize $\bm{\uptheta}_{\circ}^{t+1}$. Based on these observations, $\bm{\uptheta}_{\circ}^{t+1}$ is defined based on knowledge transfer \cite{Reptile} by clipping between $\bm{\uptheta}_{\circ}^{t}$ and $\bm{\uptheta}_{\bullet}^{t}$ as follows:

\begin{equation}
\centering
\bm{\uptheta}_{\circ}^{t+1} =\, \bm{\uptheta}_{\circ}^{t} + (k^{t})^{-\frac{1}{2}} \big(\bm{\uptheta}_{\bullet}^{t} - \bm{\uptheta}_{\circ}^{t}\big),\,\,\,\,\, \splitcond{if\,\,\emph{t}\,$\geqslant$1}.
\label{eq:ILcommonInit}
\end{equation}
\normalsize

\begin{figure}[!t]
\centering
\includegraphics[scale=0.365]{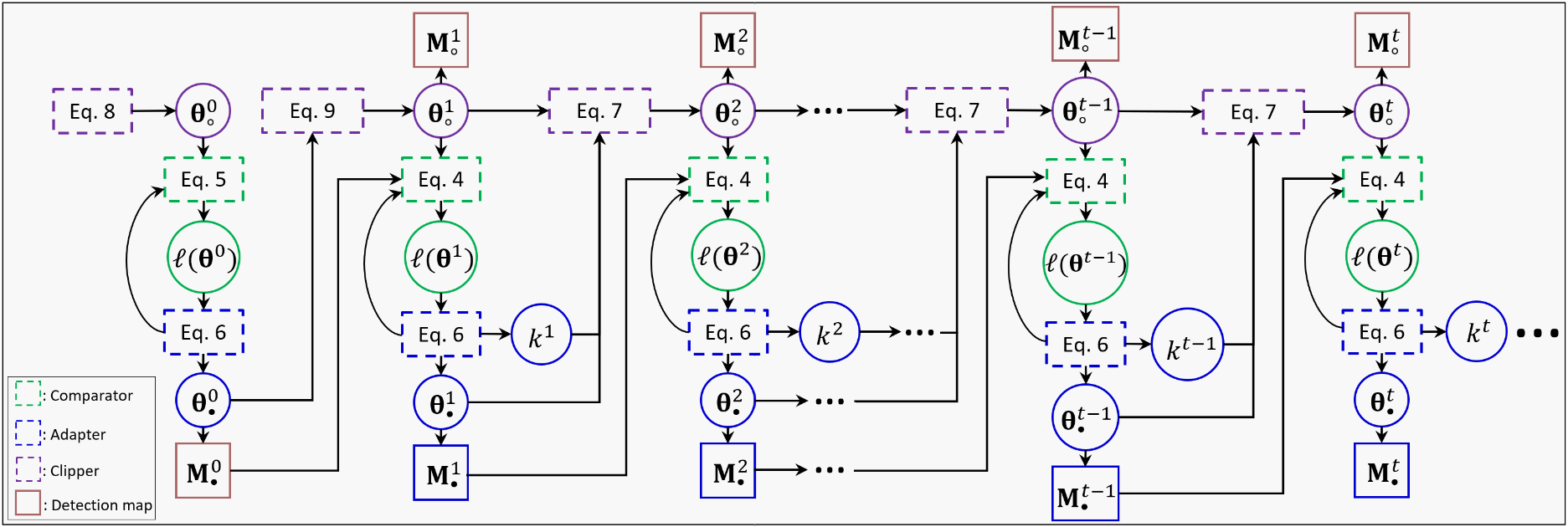}
\caption{\small{The mathematical operations of the IL along a video stream.}}
\label{fig:ILallTime}
\vspace{-5pt}
\end{figure}

For a large $k^{t}$, potentially caused by a drastic adjacent shift, $\bm{\uptheta}_{\circ}^{t+1}$ is set to be close to $\bm{\uptheta}_{\circ}^{t}$, thus rejecting the learned knowledge $\bm{\uptheta}_{\bullet}^{t}$ acquired from a potential abnormal frame $\mathbf{I}^{t}$. This is illustrated in \cref{fig:ILoneTime} (bottom-right). 

Since there is no previously learned knowledge at $t$\,=\,$0$, the MLP is randomly-initialized for this first frame as:

\begin{equation}
\centering
\bm{\uptheta}_{\circ}^{\Scale[0.65]{0}} = \widetilde{\bm{\uptheta}},
\label{eq:ILstartInitFirst}
\end{equation}
\normalsize
where $\widetilde{\bm{\uptheta}}$ indicates a random set of parameters. At $t$\,=\,$1$, the MLP is initialized without clipping parameters to avoid using the randomly-initialized parameters $\bm{\uptheta}_{\circ}^{\Scale[0.65]{0}}$ as:

\begin{equation}
\centering
\bm{\uptheta}_{\circ}^{\Scale[0.65]{1}} = \bm{\uptheta}_{\bullet}^{\Scale[0.65]{0}}.
\label{eq:ILstartInitSecond}
\end{equation}
\normalsize

\cref{fig:ILallTime} illustrates the complete functionality of the IL along a video stream, while \cref{alg:IL} summarizes it.

\begin{table}[!t]
\vspace{-17pt}
\begin{minipage}[t]{0.460\linewidth}
\centering
\begin{algorithm}[H]
\scriptsize
\caption{The IL}
\label{alg:IL}
\begin{algorithmic}[1]
\State $i \leftarrow 0$
\State $\bm{\uptheta}_{\Scale[0.65]{0}}^{t} \leftarrow \bm{\uptheta}_{\circ}^{t}$
\While{\emph{True}}
    \State $\mathbf{M}_{i}^{t} \leftarrow \big(\mathbf{I}^{t}-{f}_{\Scale[0.55]{\bm{\uptheta}}_{\Scale[0.45]{i}}^{\Scale[0.45]{t}}}(\mathbf{T}^{t})\big)^{\odot^{2}}$
    \State $\Scale[1]{\mathrm{\upepsilon}}_{i}^{t} \leftarrow \frac{1}{hw}{\big\Vert \mathbf{M}_{i}^{t} \big\Vert}_{1}$
    \State $\mathbf{M}_{\circ}^{t} \leftarrow \mathbf{M}_{\Scale[0.65]{0}}^{t}$, $\Scale[1]{\upepsilon}_{\circ}^{t} \leftarrow \Scale[1]{\upepsilon}_{\Scale[0.65]{0}}^{t}$ \textbf{when} $i=0$
    \State $\Scale[1.2]{\ell}(\Scale[1]{\bm{\uptheta}}_{i}^{t}) \leftarrow \mathcal{COMPARE}(\overline{\Scale[1]{\upepsilon}}, \Scale[1]{\mathrm{\upepsilon}}_{\bullet}^{t-1}, \Scale[1]{\mathrm{\upepsilon}}_{i}^{t})$
    \If{$\Scale[1.2]{\ell}(\Scale[1]{\bm{\uptheta}}_{i}^{t})$\,$\leqslant$\,$\overline{\Scale[1]{l}}$\,\,\textbf{or}\,\,$i$\,$\geqslant$\,$\overline{\Scale[1]{k}}$}
        \Break
    \Else
        \State $\Scale[1]{\bm{\uptheta}}_{i+1}^{t} \leftarrow \mathcal{ADAPT}(\upalpha, \Scale[1]{\nabla}_{\Scale[0.55]{\bm{\uptheta}}_{\Scale[0.45]{i}}^{\Scale[0.45]{t}}}\Scale[1.2]{\ell}(\Scale[1]{\bm{\uptheta}}_{i}^{t}))$
        \State $i \leftarrow i+1$
    \EndIf
\EndWhile
\State $\bm{\uptheta}_{\bullet}^{t} \leftarrow \bm{\uptheta}_{i}^{t}$, $\mathbf{M}_{\bullet}^{t} \leftarrow \mathbf{M}_{i}^{t}$, $\Scale[1]{\upepsilon}_{\bullet}^{t} \leftarrow \Scale[1]{\upepsilon}_{i}^{t}$
\State $\bm{\uptheta}_{\circ}^{t+1} \leftarrow \mathcal{CLIP}(\bm{\uptheta}_{\circ}^{t}, \bm{\uptheta}_{\bullet}^{t}, k^{t-1})$
\State $k^{t} \leftarrow$ $($\textbf{if} $i=0$ \textbf{then} $1$ \textbf{else} $i$ \textbf{end}$)$
\State $\mathbf{M}_{\mathrm{det}}^{t} \leftarrow$ $($\textbf{if} $t=0$ \textbf{then} $\mathbf{M}_{\bullet}^{t}$ \textbf{else} $\mathbf{M}_{\circ}^{t}$ \textbf{end}$)$
\State $\mathcal{RETURN}(\bm{\uptheta}_{\circ}^{t+1}, \Scale[1]{\upepsilon}_{\bullet}^{t}, k^{t}, \mathbf{M}_{\mathrm{det}}^{t})$
\end{algorithmic}
\end{algorithm}
\normalsize
\end{minipage}
\hfill
\begin{minipage}[t]{0.530\linewidth}
\centering
\begin{algorithm}[H]
\scriptsize
\caption{Computation of detection maps}
\label{alg:Model}
\begin{algorithmic}[1]
\State $\mathbb{M}_{\mathrm{det}} \leftarrow \emptyset$
\State $t \leftarrow 0$
\State $\bm{\uptheta}_{\circ}^{\Scale[0.65]{0}} \leftarrow \mathcal{RAND}()$
\State $\Scale[1]{\upepsilon}_{\bullet}^{\Scale[0.65]{-1}} \leftarrow \emph{null}$
\State $k^{\Scale[0.65]{-1}} \leftarrow \emph{null}$
\While{\emph{True}}
    \If{$\mathbf{V}^{t}$\emph{not exists}}
    \Break
    \EndIf
    \State $\Scale[1]{\mathbf{V}^{t}} \leftarrow \Scale[1]{\mathcal{TRANSFORM}}(\Scale[1]{\mathbf{V}^{t}})$
    \State $\Scale[1]{\mathbf{C}^{t}} \leftarrow \Scale[1]{\mathcal{TRANSFORM}}(\Scale[1]{\mathbf{C}^{t}})$
    \State ${\Scale[1]{\mathbf{A}_{1}^{t}}, \Scale[1]{\mathbf{B}_{1}^{t}}, \Scale[1]{\mathbf{A}_{2}^{t}}, \Scale[1]{\mathbf{B}_{2}^{t}}} \leftarrow \Scale[1]{\mathcal{DWT}}(\Scale[1]{\mathbf{V}^{t}})$
    \State $\mathbf{T}^{t} \leftarrow \left[\mathbf{C}^{t}, \,[\mathbf{A}_{1}^{t}]_{a_{1},:,:}, \,\mathbf{B}_{1}^{t}, \,\mathbf{B}_{2}^{t}\right]$
    \State $\bm{\uptheta}_{\circ}^{\Scale[0.6]{t+1}}, \Scale[1]{\upepsilon}_{\bullet}^{t}, k^{t}, \mathbf{M}_{\mathrm{det}}^{t} \leftarrow \mathcal{IL}(\mathbf{I}^{t}, \mathbf{T}^{t}, \bm{\uptheta}_{\circ}^{t}, 
    \Scale[1]{\upepsilon}_{\bullet}^{\Scale[0.6]{t-1}}, k^{\Scale[0.6]{t-1}})$
    \State $\mathbb{M}_{\mathrm{det}} \leftarrow \mathbb{M}_{\mathrm{det}} \bigcup \,\{\mathbf{M}_{\mathrm{det}}^{t}\}$
    \State $t \leftarrow t+1$
\EndWhile
\end{algorithmic}
\end{algorithm}
\normalsize
\end{minipage}
\vspace{-10pt}
\end{table}

\noindent \textbf{Detection and anomaly inference.}
As depicted in \cref{fig:ILallTime}, at the current timestep $t$, the MLP with initial parameters $\bm{\uptheta}_{\circ}^{t}$ results in an error map $\mathbf{M}_{\circ}^{t}$ as the detection results, hereinafter called detection maps. For all observed frames, these detection maps are:

\begin{equation}
\centering
\mathbb{M}_{\mathrm{det}} = \{\mathbf{M}_{\bullet}^{\Scale[0.65]{0}}, \mathbf{M}_{\circ}^{\Scale[0.65]{1}}, \mathbf{M}_{\circ}^{\Scale[0.65]{2}}, ..., \mathbf{M}_{\circ}^{\Scale[0.65]{t}}\},
\label{eq:ErrMaps}
\end{equation}
\normalsize
where at $t$\,=\,$0$, we use the detection map $\mathbf{M}_{\bullet}^{\Scale[0.65]{0}}$ computed after the optimization.\footnote{\scriptsize{Because $\mathbf{M}_{\circ}^{\Scale[0.65]{0}}$ is a noisy error map generated by the MLP under random-initialization.}} The detection maps can be visually displayed as heatmaps that depict the abnormal pixels,\footnote{\scriptsize{See examples of detection maps in \cref{fig:DetMaps}.}} or alternately, one can calculate the MSE values associated with these maps to numerically quantify the anomalies. \cref{alg:Model} summarizes the process to compute the detection maps.

\section{Experiments}
\noindent \textbf{Datasets.}
We test our model on three benchmark datasets. The UCSD Ped2 \cite{UCSD} dataset, which is a single-scene dataset depicting a pedestrian walkway where the anomalous events are individuals cycling, driving or skateboarding. The CUHK Avenue \cite{150FPS} dataset, which is also a single-scene dataset depicting a subway entrance with various types of anomalous events, such as individuals throwing papers and dancing. And the ShanghaiTech \cite{TSC} dataset, which is more difficult to analyze as it has complex anomalous events across multiple scenes.\footnote{\scriptsize{The dataset has \emph{13} scenes but there are no test videos for the last scene, thus we only focus on the first \emph{12} scenes.}}

\noindent \textbf{Non-continuous videos.}
In each benchmark dataset, videos depicting the same scene may not be continuously shot by one camera. Hence, our model regards these videos as different video streams. However, since such video streams may share common information, e.g., the background information, instead of randomly initializing our model on the first frame of these video streams, as specified in \cref{eq:ILstartInitFirst}, our model initializes the parameters by transferring the knowledge learned on the last frame of the previous video stream. Note that for continuous videos depicting the same scene and shot by one camera, our model computes the initial parameters as specified by \cref{fig:ILallTime}.

\noindent \textbf{Implementation details.}
All frames are transformed into gray-scale images and respectively re-sized to \emph{230}$\times$\emph{410} and \emph{240}$\times$\emph{428} for the CUHK Avenue and ShanghaiTech datasets. All pixel coordinates and pixel values are re-scaled to the range [-\emph{0.5}, \emph{0.5}]. \emph{Daubechies 2} is used as the filter for the temporal DWT on each set of video frames $\mathbf{V}^t$ of length $n$\,=\,$16$. We randomly initialize the parameters of our MLP based on the settings provided in \cite{Sin}.\footnote{\scriptsize{All activation functions are \emph{Sine} and only the last layer has no activation function.}} Adam \cite{Adam} is used with learning rates $\emph{1}$$\,\times$$\emph{10}^\emph{-4}$ and $\emph{1}$$\,\times$$\emph{10}^\emph{-5}$, respectively, on the first frame and subsequent frames of all videos. The two user-defined parameters, $\overline{\Scale[1]{\upepsilon}}$ and $\overline{\Scale[1]{l}}$, are respectively set to $\emph{1}$$\,\times$$\emph{10}^\emph{-4}$ and $\emph{1}$$\,\times$$\emph{10}^\emph{-6}$ on all videos. $\overline{\Scale[1]{k}}$ is set to \emph{500} on the first \emph{5} frames of each video, and then to \emph{100} on the remaining frames.

\noindent \textbf{Evaluation metrics.}
We use the frame-level Area Under the Curve (AUC) of the Receiver Operating Characteristic (ROC) curve as the evaluation metric. Higher AUC values indicate better model performance.

\subsection{Performance}
In \cref{tab:Performances}, we tabulate the performance of our model and other baseline models, where we highlight the best performing model in \textbf{bold} and \underline{underline} the second best performing model. The left part of \cref{tab:Performances} tabulates the performance of online VAD solutions, while the right part compares ours with offline VAD solutions. Note that since the other three online VAD solutions cannot work on videos with a large number of abnormal frames while our model does not have that restriction, substantially more videos are tested by our solution. Despite more videos being analyzed, our model outperforms the best model previously reported in \cite{MC2ST} by \emph{9.0}\% and \emph{5.8}\% AUC, respectively, on the UCSD Ped2 and CUHK Avenue datasets. Note that we report the first result computed by an online VAD solution on the ShanghaiTech dataset, i.e., an AUC value of \emph{83.1}\%. Compared to offline VAD solutions, the performance of our solution is very competitive, outperforming most of the models on the ShanghaiTech dataset with an AUC value of \emph{83.1}\%, which confirms our robustness to deal with complex scenes. The best performing offline solution is \cite{SSMTL}, which relies on multiple tasks where several networks are separately trained offline to jointly detect anomalies. Our solution attains competitive results by detecting video anomalies on-the-fly without offline training under a single framework based on frame reconstruction.

\begin{table}[!t]
\caption{\small{Comparisons in frame-level AUC values (\%) on three benchmark datasets with online VAD models (left) and offline VAD models (right).}}
\vspace{10pt}
\begin{minipage}[t]{0.362\linewidth}
\centering
\scriptsize
\begin{tabular}{ccccc}
\toprule
&
\thead[c]{\Scale[0.85]{\text{Giorno}} \\ \Scale[0.8]{\text{\cite{Shuffle}}}} & \thead[c]{\Scale[0.85]{\text{Iones}}- \\ \Scale[0.9]{\text{cu}}\,\Scale[0.8]{\text{\cite{UnMasking}}}} & \thead[c]{\Scale[0.85]{\text{Liu}} \\ \Scale[0.8]{\text{\cite{MC2ST}}}} & \thead[c]{\Scale[0.85]{\text{Ours}}} \\
\midrule
\makecell{\Scale[1]{\text{UCSD}}\\\Scale[1]{\text{Ped2}}} & - & 82.2 & \underline{87.5} & \textbf{96.5} \\ \midrule
\makecell{\Scale[1]{\text{CUHK}}\\\Scale[1]{\text{Avenue}}} & 78.3 & 80.6 & \underline{84.4} & \textbf{90.2} \\ \midrule
\makecell{\Scale[1]{\text{Shang}}-\\\Scale[1]{\text{haiTech}}} & - & - & - & \textbf{83.1} \\
\bottomrule
\end{tabular}
\normalsize
\end{minipage}
\hfill
\begin{minipage}[t]{0.628\linewidth}
\centering
\scriptsize
\begin{tabular}{cccccccccc}
\toprule
&
\thead[c]{\Scale[0.85]{\text{Chang}} \\ \Scale[0.8]{\text{\cite{Clustering}}}} &
\thead[c]{\Scale[0.85]{\text{Lu}} \\ \Scale[0.8]{\text{\cite{r-GAN}}}} & \thead[c]{\Scale[0.85]{\text{Wang}} \\ \Scale[0.8]{\text{\cite{Contrast}}}} & \thead[c]{\Scale[0.85]{\text{Sun}} \\ \Scale[0.8]{\text{\cite{SACR}}}} &
\thead[c]{\Scale[0.85]{\text{Cai}} \\ \Scale[0.8]{\text{\cite{AMMC-Net}}}} &
\thead[c]{\Scale[0.85]{\text{Lv}} \\ \Scale[0.8]{\text{\cite{MPN}}}} &
\thead[c]{\Scale[0.85]{\text{George}}- \\ \Scale[0.9]{\text{scu}}\,\Scale[0.8]{\text{\cite{SSMTL}}}} &
\thead[c]{\Scale[0.85]{\text{Liu}} \\ \Scale[0.8]{\text{\cite{HF-VAD}}}} &
\thead[c]{\Scale[0.85]{\text{Ours}}} \\
\midrule
\makecell{\Scale[1]{\text{UCSD}}\\\Scale[1]{\text{Ped2}}} & 96.5 & 96.2 & - & - & 96.6 & 96.9 & \textbf{99.8} & \underline{99.3} & 96.5 \\ \midrule
\makecell{\Scale[1]{\text{CUHK}}\\\Scale[1]{\text{Avenue}}} & 86.0 & 85.8 & 87.0 & 89.6 & 86.6 & 89.5 & \textbf{92.8} & \underline{91.1} & 90.2 \\ \midrule
\makecell{\Scale[1]{\text{Shang}}-\\\Scale[1]{\text{haiTech}}} & 73.3 & 77.9 & 79.3 & 74.7 & 73.7 & 73.8 & \textbf{90.2} & 76.2 & \underline{83.1} \\
\bottomrule
\end{tabular}
\normalsize
\end{minipage}
\label{tab:Performances}
\end{table}

\subsection{Further Studies} \label{sec:FurSty}
\begin{figure}[!t]
\centering
\includegraphics[scale=0.376]{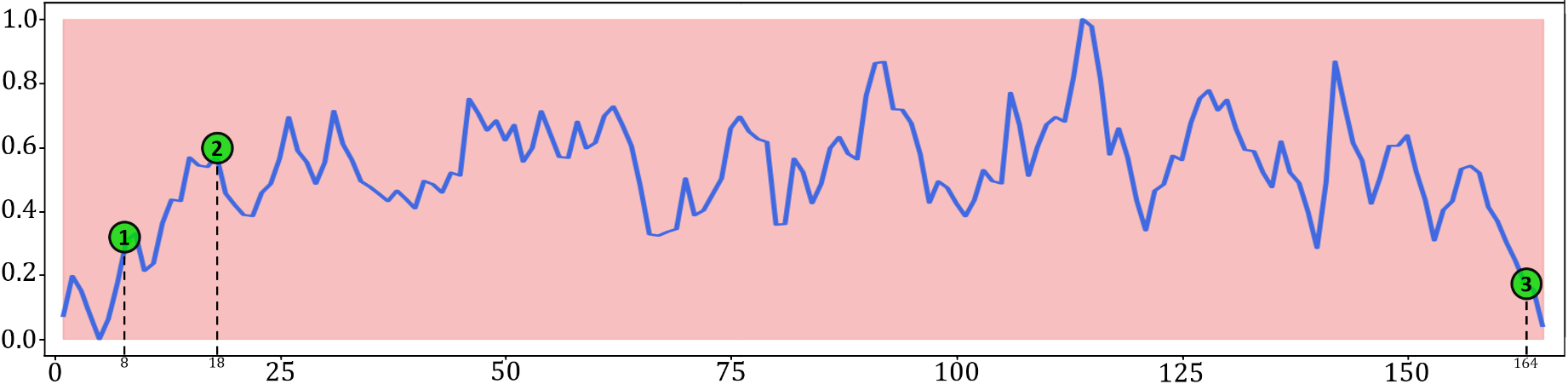}
\caption{\small{Anomaly values of the UCSD Ped2 video \emph{008}, where frames \emph{23}, \emph{33} and \emph{179} are pinned with a green circle.}}
\label{fig:FullAnom1}
\end{figure}

\noindent \textbf{Abnormal start.}
As specified in \cref{eq:ILstartLoss}, our solution optimizes the MLP on the first frame of a video stream for accurate reconstruction results even if that frame may be abnormal. However, even in such a case, the detection of anomalies at subsequent abnormal frames is not affected. We illustrate this in \cref{fig:FullAnom1}, which depicts the anomaly values computed on the UCSD Ped2 test video \emph{008} as normalized MSE values (see the blue curve). In this sequence, all frames are abnormal. Notice that even after optimizing the MLP on the first abnormal frame to achieve a low reconstruction error, which leads to an anomaly value close to \emph{0}, the anomaly values increase rather than remaining low for subsequent frames. These values eventually become quite informative to detect anomalous events. The reason for this behavior is that our solution exploits the drastic adjacent shifts that exist between abnormal frames. Hence, despite being fine-tuned on the first abnormal frame, our MLP still leads to high MSE values on subsequent abnormal frames.

\noindent \textbf{Unlimited amount of anomalies.}
\cref{fig:FullAnom1} also confirms that our solution works very well even on a video containing only abnormal frames, which is a major improvement from the previous online VAD solutions \cite{Shuffle,UnMasking,MC2ST}, whose performance degrades on videos with a large number of abnormal frames (i.e., more than \emph{50}\% of the frames depicting anomalous events). As mentioned before, a large number of abnormal frames in a video violate their assumptions that a video only contains a few abnormal frames, i.e., the spatio-temporal distinctiveness of abnormal frames in a video. Conversely, our solution makes no such assumptions, thus being capable of working on videos containing plenty of abnormal frames, or even on videos with abnormal frames only.

\begin{table}[!t]
\begin{minipage}[!b]{0.260\linewidth}
\centering
\scriptsize
\vspace{-10pt}
\captionof{table}{\small{Frame-level AUC values (\%) when testing with and without the clipper.}}
\vspace{13pt}
\begin{tabular}{ccc}
\toprule
&
\thead[c]{\Scale[0.85]{\text{Ours}} \\ \Scale[0.80]{\text{w/o\,\,clipper}}} & 
\thead[c]{\Scale[0.85]{\text{Ours}}} \\
\midrule
\makecell{\Scale[1]{\text{CUHK}}\\\Scale[1]{\text{Avenue}}} & 86.3 & 90.2 \\ \midrule
\makecell{\Scale[1]{\text{Shang}}-\\\Scale[1]{\text{haiTech}}} & 79.8 & 83.1 \\
\bottomrule
\end{tabular}
\label{tab:ClipperAnalysis}
\end{minipage}
\hfill
\begin{minipage}[!b]{0.710\linewidth}
\centering
\includegraphics[scale=0.359]{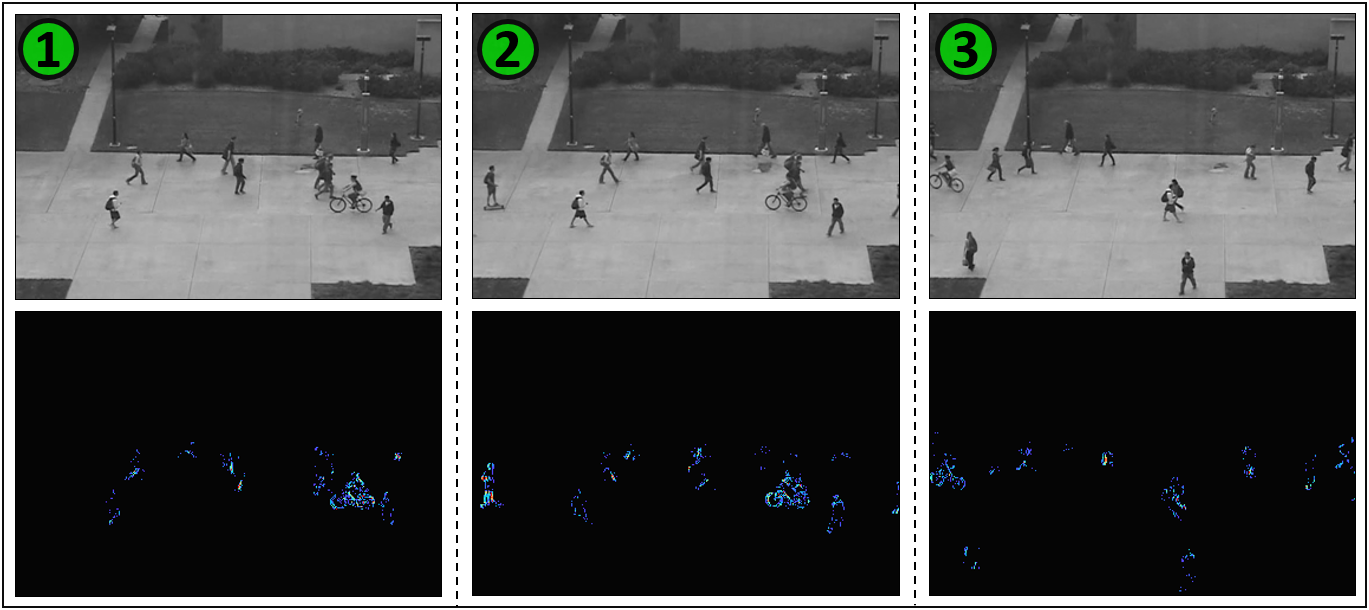}
\vspace{0.01pt}
\captionof{figure}{\small{Frame \emph{23}, \emph{33} and \emph{179} of the UCSD Ped2 video \emph{008} and their corresponding detection maps.}}
\label{fig:FullAnom2}
\end{minipage}
\vspace{-10pt}
\end{table}
\normalsize

\begin{figure}[!t]
\centering
\subfloat{\includegraphics[scale=0.423]{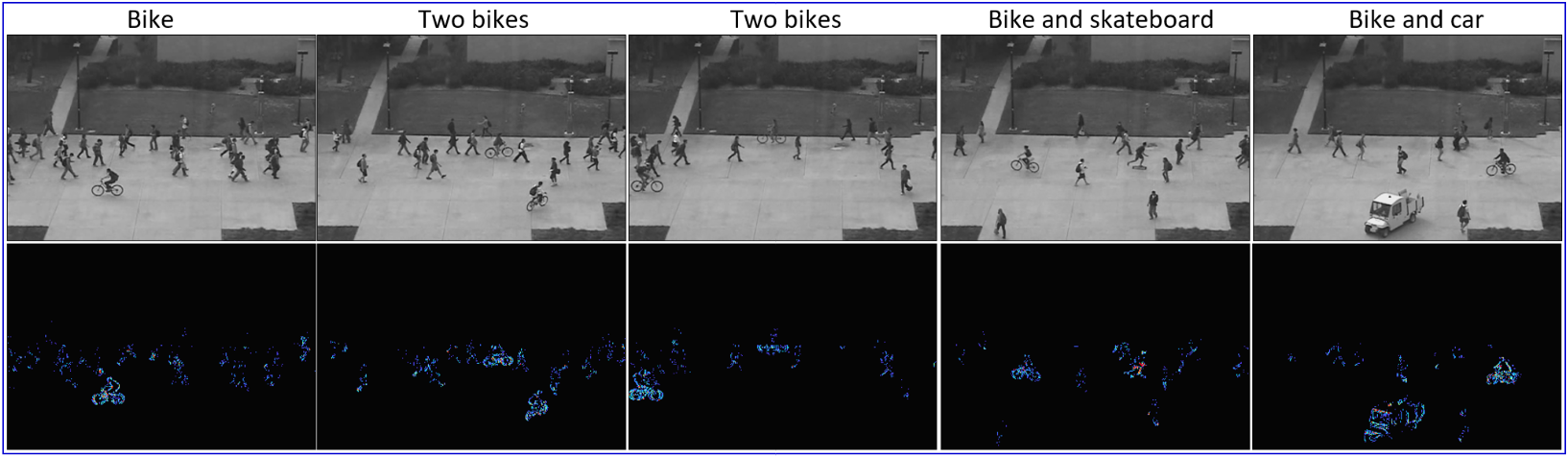}}
\hfill
\vspace{-10pt}
\subfloat{\includegraphics[scale=0.423]{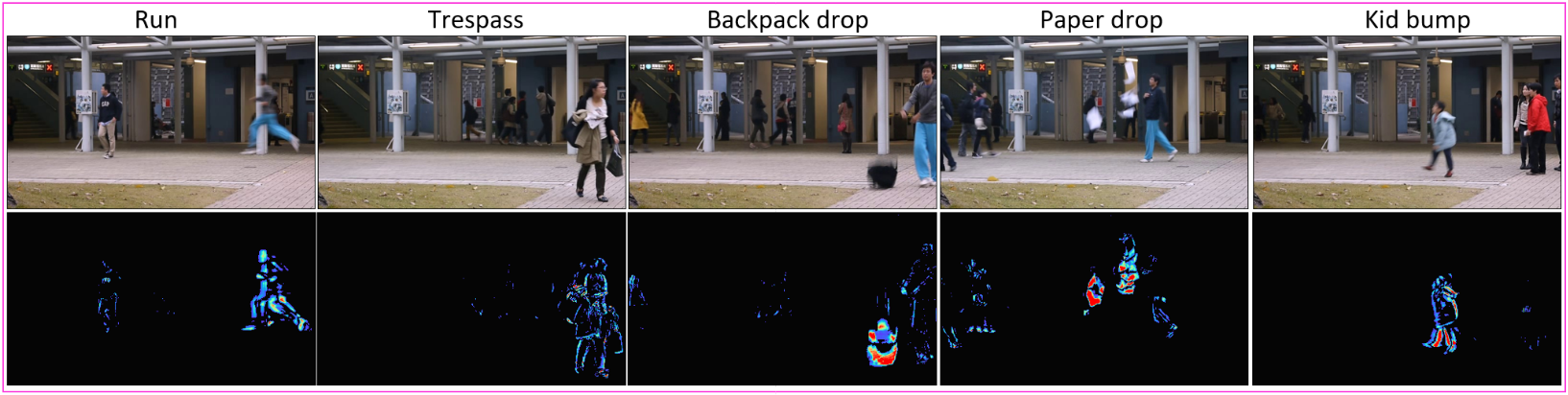}}
\hfill
\vspace{-10pt}
\subfloat{\includegraphics[scale=0.423]{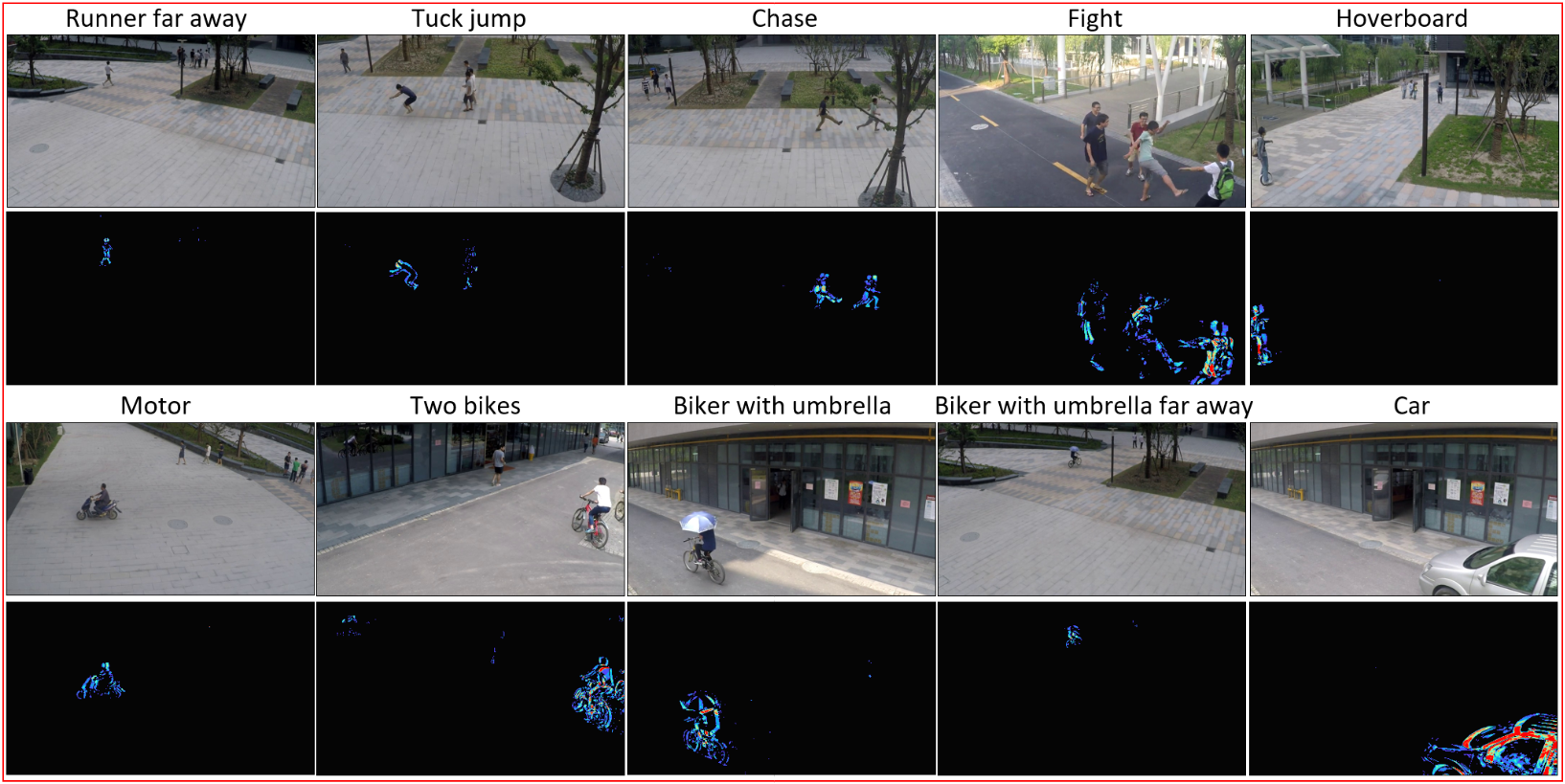}}
\caption{\small{Examples of frames and their detection maps of the UCSD Ped2 (rows \emph{1}-\emph{2}), CUHK Avenue (rows \emph{3}-\emph{4}) and ShanghaiTech (rows \emph{5}-\emph{8}) datasets, where the type of anomalous event is indicated above each example. Our solution accurately detects various types of anomalous events at the pixel-level.}}
\label{fig:DetMaps}
\end{figure}

\noindent \textbf{Pixel-level detection.}
In \cref{fig:FullAnom1}, we select three frames, i.e., frames \emph{23}, \emph{33} and \emph{179} (pinned with green circles),\footnote{\scriptsize{Index numbers in the figure are \emph{8}, \emph{18} and \emph{164}, respectively, since our model begins analyzing this sequence at the $\mathit{16}_{\mathrm{th}}$ frame.}} and illustrate these frames and their detection maps in \cref{fig:FullAnom2}. By examining these detection maps, one can see that since the cycling event is detected in all three frames, the skateboarding event should be responsible for the changes in anomaly values in \cref{fig:FullAnom1}. Specifically, its presence and disappearance, respectively, lead to an increase in the anomaly value of the second frame (frame \emph{33}) and a decrease in the third frame (frame \emph{179}). Moreover, when the skateboarder gradually enters (leaves) the scene, our solution detects it with an increasing (decreasing) trend on anomaly values (see the blue curve before frame \emph{33}, and before frame \emph{179} in \cref{fig:FullAnom1}), which shows that our solution can detect anomalous events at the frame boundaries where the anomalies are usually not evident enough.

\cref{fig:DetMaps} shows examples of detection maps and their corresponding frames\footnote{\scriptsize{We use \emph{matplotlib rainbow} as the colormap, with its colors being replaced by black at low values to increase the contrast.}}. These sample results confirm the advantages of our pixel-level detections by demonstrating that our solution can identify video anomalies at a fine granularity level. For example, it can detect the umbrellas held by bikers (see the example in the last row, third column), small abnormal objects located in the background (see the examples at the sixth row, first column, and in the last row, fourth column). The visual results in \cref{fig:DetMaps} also show that our solution accurately reconstructs the scene background and foreground where pixel errors may be hardly recognizable. This demonstrates that the IL used by our solution successfully transfers common information along the video stream.

\begin{figure}[!t]
\begin{minipage}[t]{0.344\linewidth}
\centering
\includegraphics[scale=0.173]{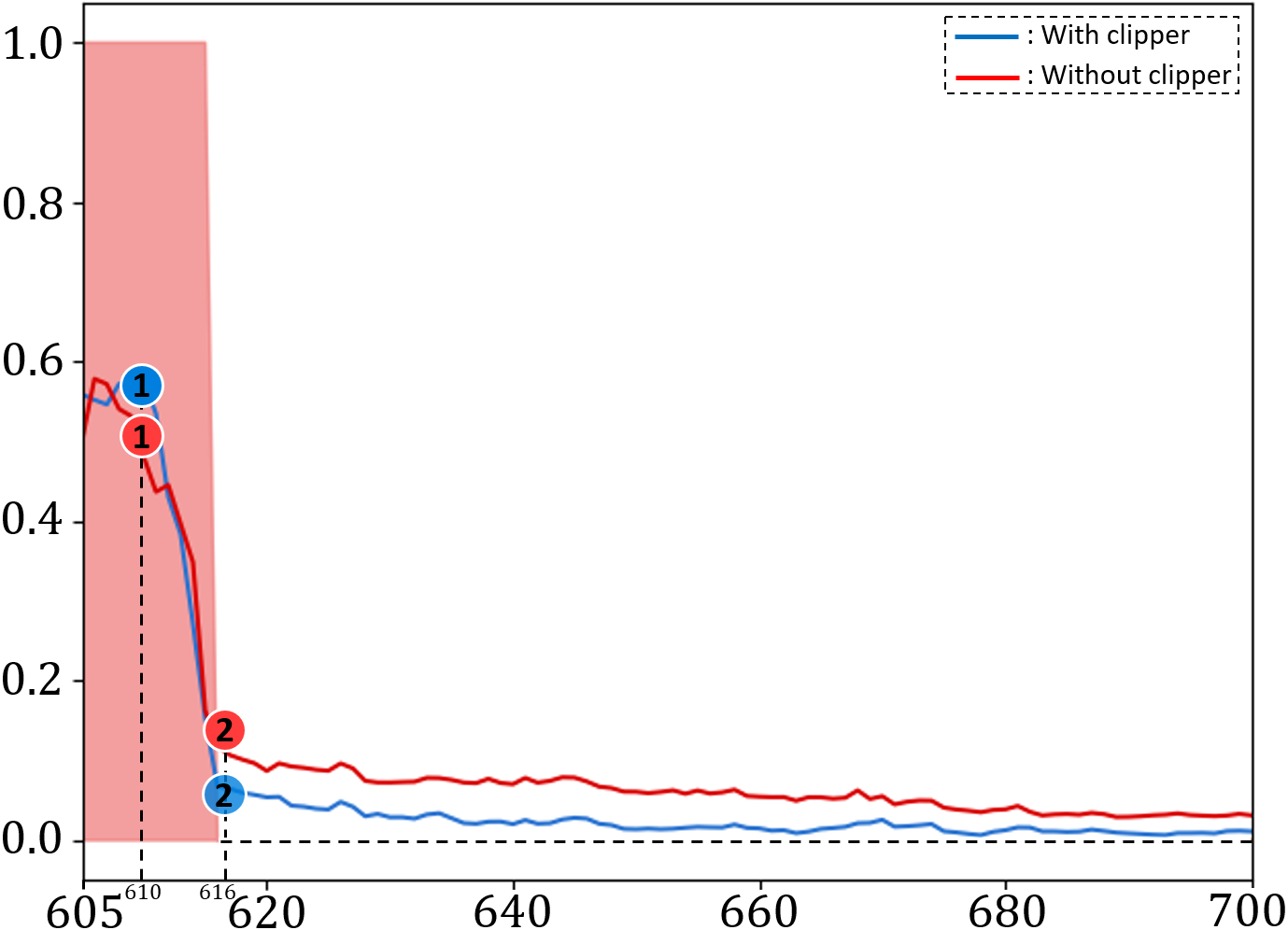}
\end{minipage}
\hfill
\begin{minipage}[t]{0.651\linewidth}
\centering
\includegraphics[scale=0.270]{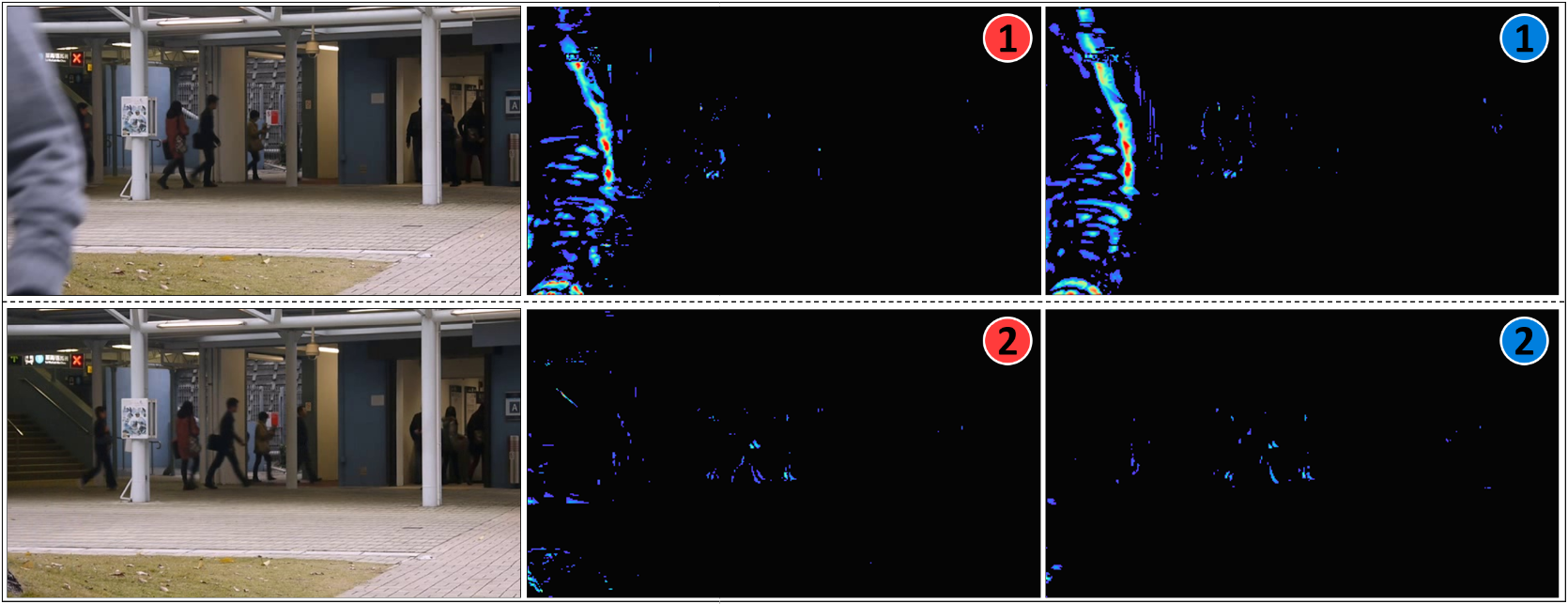}
\end{minipage}
\caption{\small{Anomaly values based on normalized MSE from frames \emph{620} to \emph{715} of the CUHK Avenue dataset video \emph{006} (left). Frame \emph{625} (row \emph{1}) and \emph{631} (row \emph{2}) and their detection maps computed without (column \emph{2}) and with (column \emph{3}) the clipper (right).}}
\label{fig:ClipperAnalysis}
\end{figure}

\noindent \textbf{The clipper.} \label{NoClip}
We perform an ablation study to showcase the functionality of the clipper. First, we use our solution on two benchmark datasets without using the clipper, i.e., directly setting $\bm{\uptheta}_{\circ}^{t} = \bm{\uptheta}_{\bullet}^{t-1}$, and then tabulate its performance in terms of AUC values with that attained by using the clipper (see \cref{tab:ClipperAnalysis}). These results show that by using the clipper, the performance increases, respectively, by \emph{3.9}\% and \emph{3.3}\% on the CUHK Avenue and ShanghaiTech datasets. 

To further understand the effect of the clipper on performance, we focus on frames \emph{620} to \emph{715} of the CUHK Avenue test video \emph{006}, which depicts the end of an anomalous event, i.e., a person trespasses and walks out of the scene. The anomaly values computed with and without the clipper are plotted in \cref{fig:ClipperAnalysis} (left) in blue and red curves, respectively. Frames \emph{625} and \emph{631} and their detection maps are depicted in \cref{fig:ClipperAnalysis} (right).\footnote{\scriptsize{Index numbers in the figure are \emph{610} and \emph{616}, respectively, since our model begins analyzing this sequence at the $\mathit{16}_{\mathrm{th}}$ frame.}} Regardless of whether the clipper is used, our solution produces informative detection maps to detect the anomalous event in frame \emph{625} (first row). For frame \emph{631} (second row), which is the first frame after the anomalous event ends, one can see that when the clipper is not used, our solution produces a noisier detection exactly in the region where the person leaves the scene. The pixels erroneously marked as abnormal in this error map are the result of overfitting the model on the previous abnormal frames. As depicted by the red curve in \cref{fig:ClipperAnalysis} (left), such overfitted model generates higher anomaly values in subsequent normal frames, which results in a poor performance. Hence, without clipping the knowledge learned from previous frames, the model may not be appropriate to be used on subsequent frames.

\section{Conclusion} \label{Conclusion}
In this paper, we proposed a solution for online VAD where offline training is no longer required. Our solution is based on a pixel-level MLP that reconstructs frames from pixel coordinates and DWT coefficients. Based on the information shifts between adjacent frames, an incremental learner is used to optimize the MLP online to produce detection results along a video stream. Our solution accurately detects anomalous events at the pixel-level, achieves strong performance on benchmark datasets, and surpasses other online VAD models by being capable to work with any number of abnormal frames in a video. Our future work focuses on improving performance by reducing the number of false positive detections.

\clearpage
%
%

\end{document}